\documentclass[10pt,twocolumn,letterpaper]{article}

\usepackage{cvpr}              

\usepackage{times}
\usepackage{epsfig}
\usepackage{graphicx}
\usepackage{amsmath}
\usepackage{amssymb}
\usepackage{booktabs}
\usepackage{bbm}
\usepackage{romannum}
\usepackage{multirow}
\usepackage{rotating}

\usepackage[pagebackref,breaklinks,colorlinks]{hyperref}

\usepackage[capitalize]{cleveref}
\crefname{section}{Sec.}{Secs.}
\Crefname{section}{Section}{Sections}
\Crefname{table}{Table}{Tables}
\crefname{table}{Tab.}{Tabs.}

\def\cvprPaperID{*****} 
\def\confName{CVPR}
\def\confYear{2022}

\begin{document}

\title{LoopDA: Constructing Self-loops to Adapt Nighttime Semantic Segmentation}

\author{Fengyi Shen\textsuperscript{1,2}, Zador Pataki\textsuperscript{2,4}, Akhil Gurram\textsuperscript{2}, Ziyuan Liu\textsuperscript{2}, He Wang\textsuperscript{3}, Alois Knoll\textsuperscript{1}\\
\noindent\textsuperscript{1}Technical University of Munich, \textsuperscript{2}Huawei Munich Research Center\\ \textsuperscript{3}Peking University, \textsuperscript{4}ETH Zurich\\
{\tt\small \textsuperscript{1}fengyi.shen@tum.de,knoll@in.tum.de,\textsuperscript{2}\{first.last\}@huawei.com,\textsuperscript{3}hewang@pku.edu.cn}
}
\maketitle

\begin{abstract}
 Due to the lack of training labels and the difficulty of annotating, dealing with adverse driving conditions such as nighttime has posed a huge challenge to the perception system of autonomous vehicles. Therefore, adapting knowledge from a labelled daytime domain to an unlabelled nighttime domain has been widely researched. In addition to labelled daytime datasets, existing nighttime datasets usually provide nighttime images with corresponding daytime reference images captured at nearby locations for reference. The key challenge is to minimize the performance gap between the two domains. In this paper, we propose LoopDA for domain adaptive nighttime semantic segmentation. It consists of self-loops that result in reconstructing the input data using predicted semantic maps, by rendering them into the encoded features. In a warm-up training stage, the self-loops comprise of an inner-loop and an outer-loop, which are responsible for intra-domain refinement and inter-domain alignment, respectively. To reduce the impact of day-night pose shifts, in the later self-training stage, we propose a co-teaching pipeline that involves an offline pseudo-supervision signal and an online reference-guided signal `DNA' (Day-Night Agreement), bringing substantial benefits to enhance nighttime segmentation. Our model outperforms prior methods on Dark Zurich and Nighttime Driving datasets for semantic segmentation. Code and pretrained models are available at {\href{https://github.com/fy-vision/LoopDA}{https://github.com/fy-vision/LoopDA}}. 
\end{abstract}

\section{Introduction}
\label{sec:intro}
Deep neural networks~\cite{rumelhart1986learning,lecun1998gradient,krizhevsky2012imagenet,lecun2015deep} have shown tremendous potential in semantic segmentation~\cite{long2015fully,chen2017deeplab,zhao2017pyramid,lin2017refinenet} tasks. However, most recent advances in this field seek to obtain higher model accuracies only based on training data under favourable viewing conditions. This is likely to hurdle the promotion of deep neural networks for applications such as visual perception for autonomous driving, where the robustness of trained models in all weather and illumination conditions is required. Models that are well trained on sunny daytime datasets fail to produce equally satisfactory results when applied to images captured in adverse conditions. Due to the lack of publicly available nighttime dataset with labels and the difficulty of creating annotations for nighttime images, semantic segmentation at nighttime remains challenging.\\
\indent In an attempt to close the performance gap, unsupervised domain adaptation (UDA) approaches~\cite{ben2007analysis,daume2009frustratingly,ganin2015unsupervised,long2016unsupervised,saito2018maximum} are becoming popular by taking advantage of labelled daytime data and obtaining models with adaptable knowledge on night domain. However, it still remains an open challenge to close the domain gap between day and night data.\\
\indent Reconstruction-based approaches~\cite{sankaranarayanan2018learning, zhu2018penalizing, chang2019all,shen2021tridentadapt} have been proven to be promising for UDA segmentation. To close the domain gap, they reconstruct either input or cross-domain translated images from the shared encoded feature map. However, in this way, the domain gap is only tackled on feature encoder level since gradient computation for the image reconstruction loss does not reach the segmentation head. Hence, we point out that segmentation outputs should also be involved in such reconstruction-based methods.\\
\indent To this point, we assume that for semantic segmentation, there is an intrinsic bidirectional connection between the input image and its segmentation output. In other words, if a pixel region in the RGB input is correctly assigned to a certain semantic class, this class distribution should in some way correspond to a specific pattern back in the input space, which leads to such semantic prediction. Therefore, when the semantic output is involved in creating an image, the resulting image should appear similar to the input image regarding the class-specific textures. Otherwise, the semantic predictions are likely to be wrong and should be further fine-tuned by a reconstruction loss. However, given that semantic outputs are only probability maps, it makes less sense to utilize them alone for image reconstruction without combining the encoded latent features. Therefore, to make the best use of the input data and the segmentation, we propose to construct a self-loop that can associate the encoded latent feature incorporating the segmentation output to the input image. Particularly, this provides clues for the segmentation refinement of unlabelled target data by learning from the self-loop of the labelled source data.\\
Additionally, it also comes to the question of how to best utilize the predicted semantic maps of the daytime reference data. Since directly applying their predicted static labels to guide nighttime segmentation will result in wrong predictions due to the view changes.\\ 
\begin{figure}[htb!]
\vspace{-5mm}
\centering
\includegraphics[width=0.75\columnwidth,clip=true,trim=0 0 0 0]{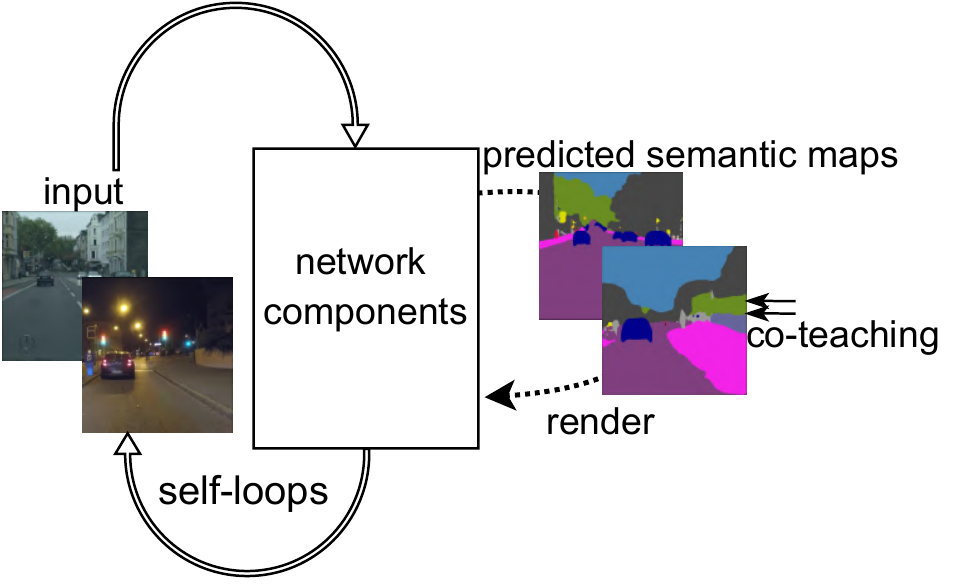}
\caption{ A systematic overview of LoopDA framework. Predicted semantic maps are refined by being rendered into the encoded features in different levels of self-loops. To condition our networks, self-loops of the daytime domain are trained with labels, thus the nighttime segmentation outputs get rectified accordingly through the self-loops. To deal with day-night pose shifts, offline and online co-teaching is performed for self-training.}
 \label{fig:overall}
\end{figure}\\
Given the aforementioned aspects, in this work we propose LoopDA, a novel framework for domain adaptive nighttime semantic segmentation. It contains two different levels of closed loops (inner and outer) that focus on reducing the domain discrepancy between day and night. Our contributions can be summarized as follows:
\begin{itemize}
\setlength{\itemsep}{0pt}
\setlength{\parskip}{0pt}
\setlength{\parsep}{0pt}
\item We introduce a LoopDA framework. It consists of self-loops where the learning of image segmentation benefits from reconstructing the input data using encoded latent features rendered by the predicted semantic maps. In a warm-up training stage, the self-loops comprise of an inner-loop responsible for intra-domain refinement, and an outer-loop taking care of inter-domain alignment to reduce the domain gap; 
\item For the self-training stage, we propose a novel pseudo-supervision strategy to allow two co-supervision signals to refine the model prediction for nighttime inputs. These include an offline signal derived from predictions of night training images, and an online signal `DNA' encouraging the identical semantic predictions of static classes between nighttime and their daytime reference image for pseudo-labelling. This tackles the pose shifts without introducing additional computation steps such as pose estimation and depth warping; 
\item Our trained models attain state-of-the-art performances on benchmark datasets for UDA semantic segmentation at nighttime.
\end{itemize}

\section{Related Work}
\label{related_work}
\subsection{Reconstruction based training}
Autoencoding~\cite{bank2020autoencoders,baldi2012autoencoders} is the basic format of unsupervised learning, which focuses on reconstructing the input data to learn its latent representations. It is a fundamental building block of many popular deep learning based architectures~\cite{long2015fully,chen2017deeplab,zhu2017unpaired,kingma2013auto,feng2020deep,park2019semantic,zhang2020cross}. Interestingly, LabelEnc~\cite{hao2020labelenc} introduces a label encoding function, mapping the ground-truth labels into latent embedding via an autoencoder, approximating the “desired” intermediate representations and acting as an auxiliary supervision to boost the object detection task. Additionally, the most recent advancement~\cite{he2021masked,wei2021masked} in the field of representation learning also indicates that autoencoding is a meaningful step for learning visual features. For instance, MAE~\cite{he2021masked} claims that having an image reconstruction pretraining stage using masked inputs can produce an effective image encoder for image classification tasks. Furthermore, MaskFeat~\cite{wei2021masked} suggests that it is also promising to reconstruct the features (e.g., HOG~\cite{dalal2005histograms}) of the masked inputs for similar tasks. Unlike from image classification, semantic segmentation task requires various representations for different pixel locations in the feature map instead of predicting only one overall class for an image. Regarding this, image reconstruction driven by the predicted label map is introduced in~\cite{yang2020label} to fine-tune the segmentation results on cross-domain data. However, predicted labels are from probability outputs and do not contain RGB information of the input data. Therefore, image reconstruction only from labels makes less sense without involving the encoded latent features. To this end, our method constructs self-loops of the input data using latent features rendered by the predicted semantic labels, thus manipulating the latent space in a more comprehensive way.
\subsection{Style transfer based training}
Since image style translation approaches~\cite{zhu2017unpaired,liu2017unsupervised,huang2018multimodal} based on GANs~\cite{goodfellow2014generative,arjovsky2017wasserstein,mao2017least,jolicoeur2018relativistic,karras2019style} can be trained in an unsupervised manner, they have been widely adopted in solving UDA problems. In~\cite{hoffman2018cycada,yue2019domain}, target-like images transferred from the source domain are used to train a segmentation model that attains better performance towards target domain data. The concept of image style translation has been well accepted in nighttime semantic segmentation. In earlier works~\cite{sun2019see,romera2019bridging,sakaridis2019guided,sakaridis2020map}, image domain translation modules are adopted either to augment nighttime into daytime styles or vice versa to better align the differences in training data appearance. However, unlike other scenarios of applying image style translation, domain transfer between day and night data is more challenging, e.g., dealing with the dark invisible regions from night images or preserving the useful semantic contents in daytime images during the domain transfer. Therefore, in our outer loop training, we propose a semantic map rendering method to better assist the day-night image transfer while enabling the nighttime semantic map to be fine-tuned. The outer loop of LoopDA is inspired by the training philosophy of CycleGAN~\cite{zhu2017unpaired}, however, we point out that CycleGAN is built for image translation while LoopDA focuses on learning domain agnostic representations from the encoder and semantic classifier, and refines nighttime predictions through semantic rendering.

\subsection{Daytime guidance for nighttime segmentation}
Given the challenging illumination condition at nighttime, it is difficult to segment dark regions such as sky, trees and buildings, etc. Therefore, it is reasonable that the segmentation is guided by daytime reference images. In~\cite{wu2021dannet}, a relighting network is first used as pre-processing module to narrow the illumination gap between all input data, and then optimize the nighttime predictions using static maps from daytime re-weighted by prediction probabilities. However, this requires the relighting network to be coupled even for post-training inference. Additionally, directly applying daytime static masks for nighttime supervision can be problematic because the pose changes lead to wrong prediction. On top of~\cite{sakaridis2019guided},~\cite{sakaridis2020map} introduces geometrical alignment by roughly estimating the camera motion between daytime reference and nighttime images, and warps daytime predictions to fit the nighttime ones using Monodepth2~\cite{godard2019digging}. This mainly tackles the pose change but the estimated poses are not accurate because of the domain gap. Moreover, it requires multiple processing steps only for geometrical alignment. To prevent the nighttime data from overfitting to daytime label guidance, we propose `DNA', a technique that gradually refines nighttime segmentation simply by considering common static predictions between daytime and nighttime as an online supervision signal, which better avoids the model to be misguided by pose-shifted daytime reference labels.

\section{Proposed Method}
\begin{figure*}[htb!]
\centering
\setlength{\belowcaptionskip}{-12pt}
\includegraphics[width=1.8\columnwidth]{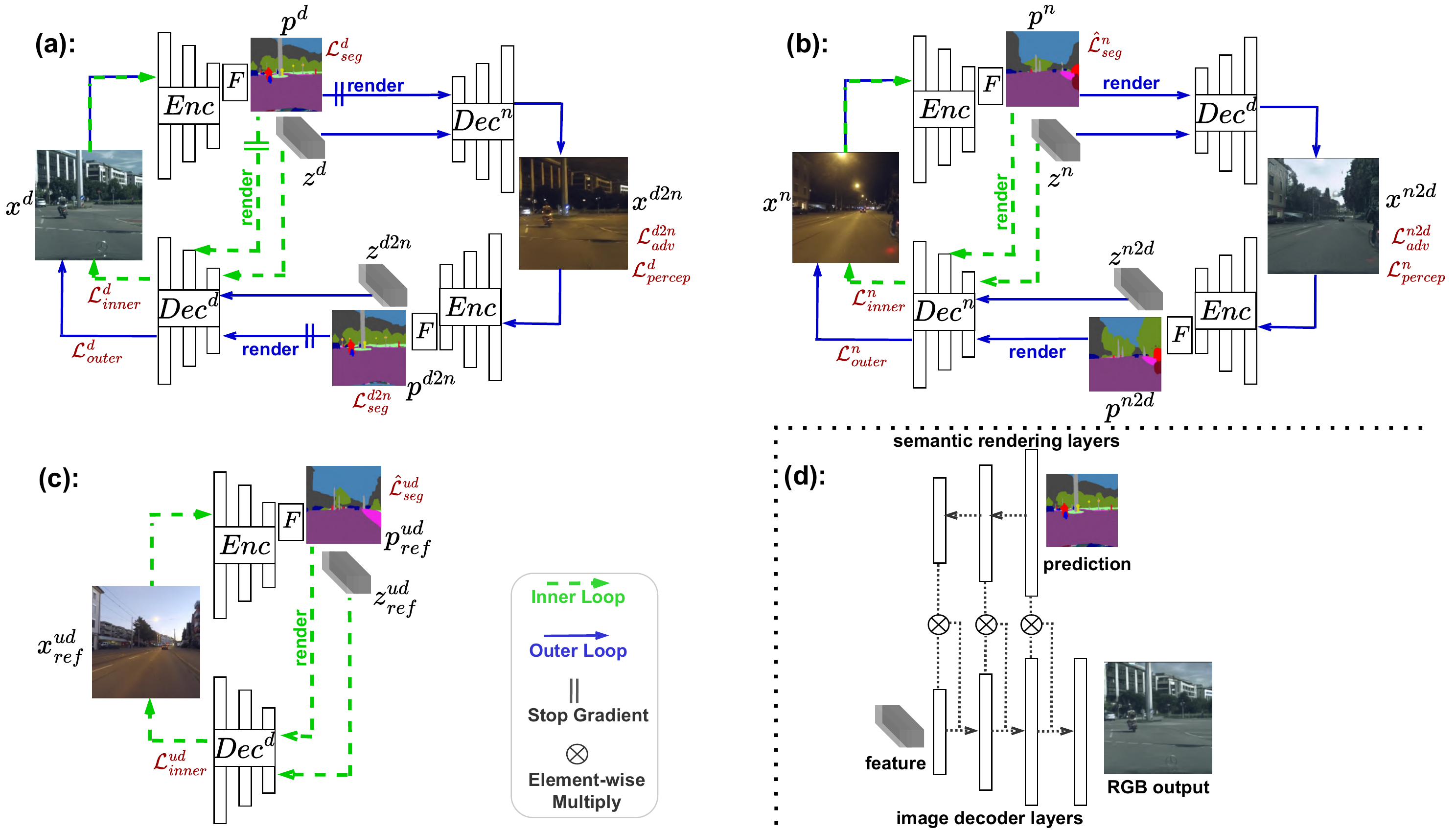}%
\caption{The scheme of training LoopDA. A self-loop starts with an input image, and loops back to the input itself.  The self-loops are constructed in two levels: intra-domain inner loop and inter-domain outer loop. With joint training on daytime and nighttime inputs, the model implicitly learns to map or refine each RGB pixel to its correct semantic assignment.}%
\label{fig:model_diagram_loopda}
\end{figure*}
Before describing our framework and learning steps, we introduce the input data for training. The training sets consist of daytime source domain images $x^d \in \mathcal{X}^d$ with corresponding per-pixel semantic annotations $y^d \in \mathcal{Y}^d$, and unlabelled images $x^n \in \mathcal{X}^n$ from the nighttime target domain. Additionally, each $x^n$ comes with an unlabelled daytime reference image $x^{ud}_{ref}$. The goal is to learn transferable semantic knowledge from the labelled daytime domain towards unlabelled nighttime domain with help of $\mathcal{X}^{ud}_{ref}$.

\subsection{Self-loops on cross-domain levels}
\label{sec:loopda}
In this section, we propose LoopDA, providing a new perspective of domain adaptation for nighttime semantic segmentation. It builds inner and outer self-loops back to the input data using feature maps rendered by semantic outputs, gradually refining nighttime segmentation results guided through self-loops with labelled daytime data. For self-training, other than the offline pseudo-label generation, we propose a novel `DNA' strategy, which takes advantage of reference daytime guidance but efficiently reduces the misguidance from pose-shifted wrong labels. Our framework mainly consists of four sub-networks: a feature encoder $Enc$, a semantic classifier $F$, a daytime decoder $Dec^d$, and a nighttime decoder $Dec^n$ incorporated with semantic rendering layers.  
\subsubsection{Loop construction for labelled daytime domain}
For the labelled daytime domain, the purpose of loop constructions is to learn a model that is able to build consistent looped mapping between an input $x^d$ and the ground-truth $y^d$ tuned by full supervision, setting the foundation for model adaptability towards nighttime domain. This involves an inner loop and an inter-domain outer loop in one training iteration. As shown in Fig.~\ref{fig:model_diagram_loopda}(a), $x^d$ is passed to $Enc$ and further to $F$ to obtain a deep latent feature $z^d = Enc(x_d)$, and a probability map $p^d = softmax(F(z^d))$, respectively. Since $y^d$ is used to supervise $p^d$, we calculate the cross-entropy loss,
\begin{align}
    \label{eq:source_seg}
    &{\mathcal{L}^{d}_{seg}} =  \sum_{h,w}\sum_{c} -{y}^{d}_{(c,h,w)} \log(p^{d})_{(c,h,w)}
\end{align}
\noindent where ${h}$, ${w}$ and $c$ are height, width and number of semantic classes, respectively.\\
Thus, any pixel in RGB space is associated with a particular semantic category in probability space.
The other way around, we want the model to learn an awareness that, if semantic predictions in $p^d$ are all correct, they can also be mapped back to the input $x^d$, reflecting different textures or patterns for the corresponding semantic classes. Given the appearance diversity of intra-class patterns, it makes less sense to recover $x^d$ simply based on $p^d$ elements that are merely probabilities. Therefore, we propose a domain-specific image decoder $Dec^d$ with semantic rendering layers where $p^d$ is incorporated into $z^d$ for image recovery. Fig.~\ref{fig:model_diagram_loopda}(d) presents details of our image decoder: $p^d$ is first encoded by several fully-convolutional layers, and the resulting semantic features are multiplied layer-wisely by the raw features acquired from $z^d$ decoder layers. Additionally, the raw features are combined into the rendered features using a skip connection following~\cite{he2016deep}, such that semantically rendered features do not override and spatial information in $z^d$ can be preserved. To close the intra-domain self-loop, $Dec^d$ is encouraged to reproduce $x^d$ constrained by the inner-loop loss for image reconstruction,
\begin{align}
    \label{eq:img_inner_d}
    \mathcal{L}^{d}_{inner} = ||{Dec^{d}(z^d, p^d)} - {x}^{d}||_{1}
\end{align}
To ensure that $p^d$ is accurate enough for rendering, we fuse $p^d$ and $y^d$ for fine-grained rectification via linear combination (details in Supplementary). Given that daytime is a labelled domain, the aim is to teach $Enc$, $F$ and $Dec^d$ for self-loop reconstruction with help of ground-truths $y^d$, therefore, it's unnecessary to let the fused $p^d$ be tuned in the self-loop, and we detach it before semantic rendering happens in $Dec^d$.\\
Meanwhile, $p^d$, together with $z^d$, goes into an inter-domain outer loop (blue path in Fig.~\ref{fig:model_diagram_loopda}(a)). Given that image translation has been known to be a meaningful component for day-night domain adaptation~\cite{dai2018dark,sun2019see}, hence, we pass $p^d$ and $z^d$ to a nighttime decoder $Dec^n$ to obtain a nighttime version of $x^d$, i.e., $x^{d2n} = Dec^{n}(z^d, p^d)$, which is learned by an adversarial loss,
\begin{align}
    \label{eq:trans_adv_d2n}
    \mathcal{L}^{d2n}_{adv} = (D^{n}({x}^{n}))^2 + (1 - D^{n}(x^{d2n}))^2
\end{align}
where ${D^{n}}$ is the nighttime domain discriminator, which is omitted from Fig.~\ref{fig:model_diagram_loopda} for simplicity. The challenging part of image translation from day to night is that clear contents such as building, trees and sky in $x^d$ can become invisible in $x^{d2n}$ as $Dec^n$ learns to mimic dark night appearance. Hence, the semantic rendering in our $Dec^n$ is able to help preserve daytime contents during translation. 
Moreover, in order to improve the perceptual quality of $x^{d2n}$, we introduce a perceptual consistency loss between $x^d$ and $x^{d2n}$ as follows,
\begin{align}
    \label{eq:PC}
    \mathcal{L}^{d}_{percep} = \lambda_{z}||z^{d2n} - z^d||_{1} + \lambda_{l}LPIPS(x^{d2n},x^d)
\end{align}
$\mathcal{L}^{d}_{percep}$ in Equation~\ref{eq:PC} places constraints from two different perspectives and comprises two parts: L1-norm semantic consistency loss between $z^d$ and $z^{d2n}$ obtained based on the shared encoder $Enc$, and LPIPS~\cite{zhang2018unreasonable} loss between $x^d$ and $x^{d2n}$ to measure structural similarity.\\
Moreover, in this inter-domain outer loop, we expect $Enc$ and $F$ to learn domain agnostic knowledge, thus treating $x^d$ and $x^{d2n}$ semantically equivalent to reduce the domain-specific bias. Therefore, we also compute supervised segmentation loss for the cross-domain semantic prediction $p^{d2n}$,
\begin{align}
    \label{eq:d2n_seg}
    &{\mathcal{L}^{d2n}_{seg}} =  \sum_{h,w}\sum_{c} -{y}^{d}_{(c,h,w)} \log(p^{d2n})_{(c,h,w)}
\end{align}
Finally, to further enforce $Enc$ and $F$ to produce domain-indistinguishable representations of input data, we encourage that $Dec^d$, taking $z^{d2n}$ and $p^{d2n}$, closes the inter-domain self-loop towards $x^d$, which means $Dec^{d}(z^d, p^d)\approx x^d \approx Dec^{d}(z^{d2n}, p^{d2n})$. This can be described by the outer loop loss,
\begin{align}
    \label{eq:img_outer_d}
    \mathcal{L}^{d}_{outer} = ||{Dec^{d}(z^{d2n}, p^{d2n})} - {x}^{d}||_{1}
\end{align}
Similar to Equation~\ref{eq:img_inner_d}, $p^{d2n}$ is fused by $y^d$ and its gradient tracking is also disabled for semantic rendering. The inner and outer loop training on the labelled daytime domain builds up a prototype for universal and robust representation between day and night, preparing the network parameters to adapt to the unlabelled nighttime data.

\subsubsection{Loop adaptation for nighttime domain}
\textbf{Unlabelled nighttime data}. Supported by the established daytime loops, in the same training iteration, we also pass $x^n$ to the shared networks for self-loop construction. Following a dual data flow of daytime domain, we first build the inner loop (see Fig.~\ref{fig:model_diagram_loopda}(b)) by inner loss $\mathcal{L}^{n}_{inner}$ similar to Equation~\ref{eq:img_inner_d}, forcing $Dec^{n}(z^n, p^n)\approx x^n$. However, instead of stopping gradient computation in semantic rendering, for the nighttime domain we always keep $p^n$ fine-tuned during loop construction. In other words, if $p^n$ is not able to help $z^n$ reconstruct $x^n$ by semantic rendering, it will get rectified towards a correct prediction based on the knowledge adapted from the daytime loops. \\
Like for it's labelled daytime counterpart, for the unlabelled nighttime domain, to enhance the domain-agnostic property of $Enc$ and $F$, we also conduct an outer loop to map $x^n$ to $x^{n2d}$ and back to $x^n$ (i.e., $x^n$ to $x^{n2d}$ to $x^n$) assisted by $Dec^{d}$ and $Dec^{n}$. Slightly different from Equation~\ref{eq:trans_adv_d2n}, in this process, the night to day translation $x^{n2d}$ is generated using an adversarial loss $\mathcal{L}^{n2d}_{adv}$ by considering not only $x^d$ but also $x^{ud}_{ref}$ as real. The reason is that, in some cases, the camera locations for $x^n$ and $x^{ud}_{ref}$ are close enough, so that some static objects in the dark regions of $x^n$ can be better recovered in $x^{n2d}$ through example guided image translation. Thus, based on a daytime domain discriminator $D^{d}$, $\mathcal{L}^{n2d}_{adv}$ is given as,
\begin{align}
    \label{eq:trans_adv_n2d}
    \mathcal{L}^{n2d}_{adv} = (D^{d}({x}^{d}))^2 + (D^{d}(x^{ud}_{ref}))^2 + (1 - D^{d}(x^{n2d}))^2
\end{align}
In addition, to preserve semantic consistency and structural similarity between $x^n$ and $x^{n2d}$, a perceptual loss $\mathcal{L}^{n}_{percep}$ is also computed following Equation~\ref{eq:PC}.\\
To close the outer self-loop, we want $Enc$ and $F$ to be invariant to any domain shift from night to day, i.e., the intermediate outputs such as $z^{n2d}$ and $p^{n2d}$ should be able to help $Dec^{n}$ recover $x^n$ in line with the inner loop, meaning $Dec^{n}(z^n, p^n)\approx x^n \approx Dec^{n}(z^{n2d}, p^{n2d})$. This is supported by the outer loop loss $\mathcal{L}^{n}_{outer}$ similar to Equation~\ref{eq:img_outer_d}. While minimizing $\mathcal{L}^{n}_{outer}$, $p^{n2d}$ is fine-tuned to produce more reasonable prediction. Most importantly, as Fig.~\ref{fig:model_diagram_loopda}(b) shows, the gradient computation of the outer loop loss $\mathcal{L}^{n}_{outer}$ can be traced back to all sub-networks in LoopDA. Hence, they are all optimized accordingly to align with the knowledge learned from the outer loop of the labelled daytime domain. With constructed inner and outer self-loops, the domain gap is gradually reduced via intra-domain refinement and inter-domain alignment.\\
\textbf{Unlabelled daytime reference data} For the attached daytime reference images $x^{ud}_{ref}$ that are unlabelled, we just build an inner loop in Fig.~\ref{fig:model_diagram_loopda}(c) with loss $\mathcal{L}^{ud}_{inner}$ to refine $p^{ud}_{ref}$ guided by the labelled daytime data $\{x^{d}, y^{d}\}$.

\subsection{Reference guided self-training}
\label{sec:self_train}
We train LoopDA adopting the popular stage-wise pipeline in domain adaptation~\cite{choi2019self,du2019ssf,kim2020learning,li2019bidirectional,pan2020unsupervised,zou2018unsupervised,zou2019confidence}, i.e., using a warm-up stage and a self-training stage. The warm-up stage is described in Sec.~\ref{sec:loopda}, on top of which, we introduce segmentation losses $\hat{\mathcal{L}}^{ud}_{seg}$ and $\hat{\mathcal{L}}^{n}_{seg}$ for unlabelled data $x^{ud}_{ref}$ and $x^{n}$ based on pseudo-labels during the self-training stage.\\
To improve the quality of daytime reference label for better guidance on nighttime domain, as Fig.~\ref{fig:model_diagram_loopda}(c) shows, we compute $\hat{\mathcal{L}}^{ud}_{seg}$ as follows,
\begin{align}
    \label{eq:ud_ref_seg}
    &{\hat{\mathcal{L}}^{ud}_{seg}} =  \sum_{h,w}\sum_{c} -\hat{y}^{ud}_{(c,h,w)} \log(p^{ud}_{ref})_{(c,h,w)}
\end{align}
where $\hat{y}^{ud}$ is acquired offline following~\cite{li2019bidirectional}.
\begin{figure}[t!]
\centering
\setlength{\belowcaptionskip}{-12pt}
\includegraphics[width=\columnwidth,clip=true,trim=0 0 0 0]{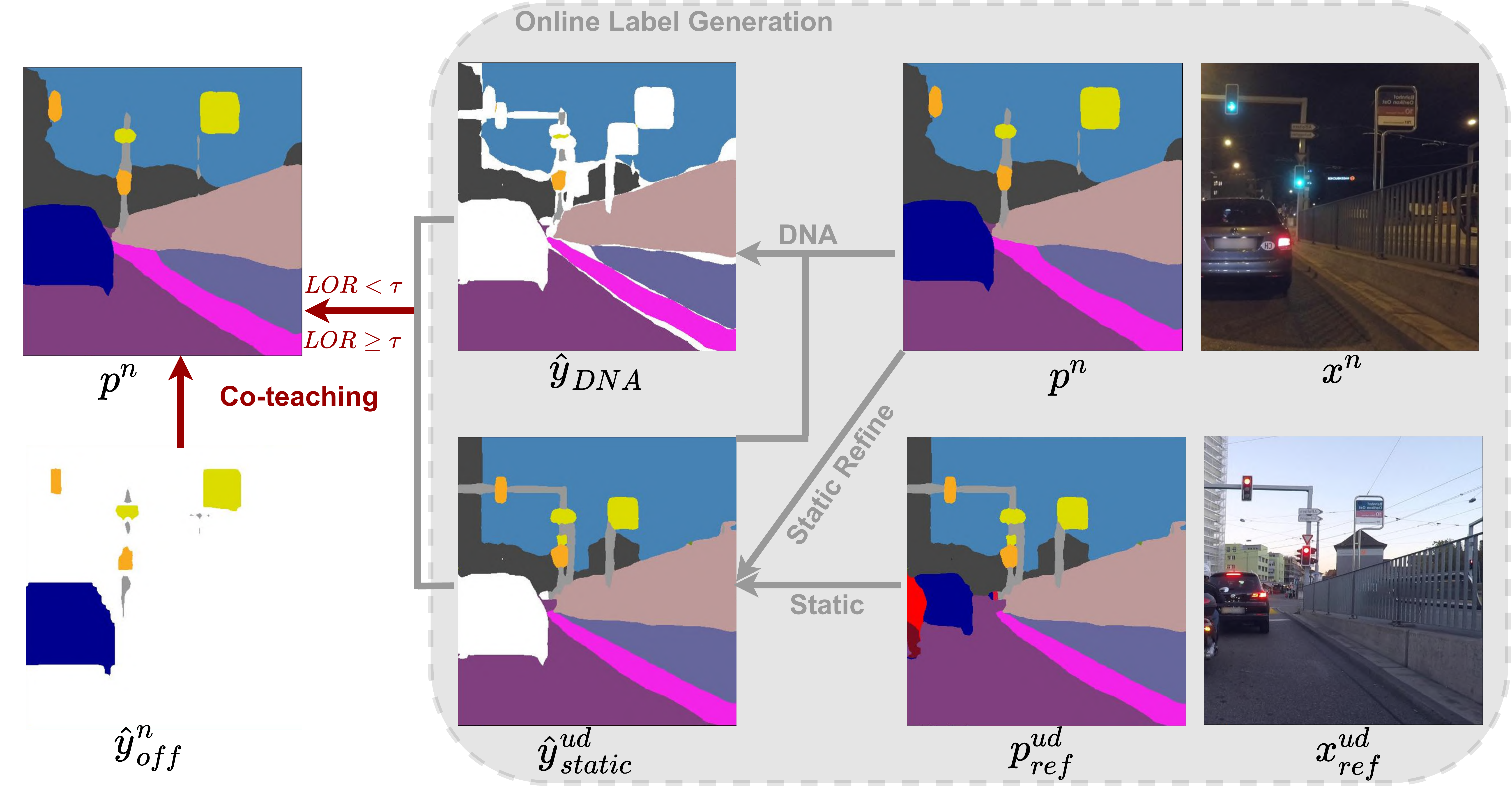}
\caption{ An illustration of the pseudo-supervision strategy in LoopDA. To prevent $p^n$ to be misguided by the pose-shifted $\hat{y}^{ud}_{static}$, the co-teaching involves an online signal based on Day-Night Agreement (DNA) of predicted labels and an offline signal from the warm-up model.}
\label{fig:self_training}
\end{figure}\\
Next, we present our proposed self-training strategy for nighttime domain. Due to the fact that most static objects reflect light weakly at night, which makes segmentation quite challenging. To tackle this issue, we propose to take the best advantage of the static labels from each daytime reference image but also prevent $p_n$ to learn from wrong labels due to pose shift. As illustrated in Fig.~\ref{fig:self_training}, given the current daytime reference prediction $p^{ud}_{static}$, we filter out all pixel locations that are associated with dynamic classes in both $p^n$ and $p^{ud}_{static}$, obtaining a static label map $\hat{y}^{ud}_{static}$ on-the-fly. In parallel, we also generate $\hat{y}_{\scriptsize DNA} = \hat{y}^{ud}_{static}\cap \mathcal{O}(\arg\max(p^n))$ online, indicating the Day-Night Agreement (DNA) in predicted static labels ($\mathcal{O}$ is one-hot operator). $\hat{y}_{\scriptsize DNA}$ is less affected by pose shifts, meanwhile, the pixel regions that are double checked by `DNA' have larger chance to be correct. A question thus rises on how to choose between $\hat{y}^{ud}_{static}$ and $\hat{y}_{\scriptsize DNA}$ as the online reference static label for $p^{n}$. Intuitively, if there is a very small camera pose change, $\hat{y}^{ud}_{static}$ is preferred, otherwise $\hat{y}_{\scriptsize DNA}$ should take the charge. We observe that semantic categories such as traffic signs, traffic lights and poles are quite sensitive to camera pose shift, which are referred to as shift sensitive classes (SSC). Following this clue, between the derived label maps $l^{n} = \arg\max(p^n)$ and $l^{ud}_{ref} = \arg\max(p^{ud}_{ref})$, we compute a label overlapping ratio (LOR) for SSC, 
\begin{align}
    \label{eq:LOR}
    {LOR} = \frac{2\sum_{i\in N}\mathbbm{1}\{l^{n}_{i} \in SSC\}\cdot\mathbbm{1}\{l^{ud}_{ref,i}\in SSC\}}{\sum_{i\in N}\mathbbm{1}\{l^{n}_{i}\in SSC\} + \mathbbm{1}\{l^{ud}_{ref,i}\in SSC\}}
\end{align}
where $N=h\cdot w$, and $\mathbbm{1}$ is an indicator checking whether the current label pixel belongs to SSC. Therefore, if LOR exceeds a certain threshold $\tau$, we are then able to confirm that there is a very small pose shift between $x^{n}$ and $x^{ud}_{ref}$, meaning that $\hat{y}^{ud}_{static}$ instead of $\hat{y}_{\scriptsize DNA}$ should be applied on $p^{n}$ as the online supervision signal, which is given by $\hat{y}^n_{on} =
\begin{cases}
\hat{y}^{ud}_{static},  & LOR\ge\tau \\
\hat{y}_{\scriptsize DNA}, & LOR<\tau
\end{cases}$. This empowers reliable static label generation without introducing additional camera pose estimation and depth warping modules. Combined with the offline label $\hat{y}^{n}_{off}$ containing dynamic classes and SSC from warm-up stage, we compute $\hat{\mathcal{L}}^{n}_{seg}$ by co-teaching of these two supervision signals as follows,
\begin{align}
    \label{eq:n_seg}
    &{\hat{\mathcal{L}}^{n}_{seg}} =  \sum_{h,w}\sum_{c} -[(\hat{y}^{n}_{off} + \hat{y}^{n}_{on}) \log(p^{n})]_{(c,h,w)}
\end{align}
So far, the mentioned losses can be summarized into five categories for all data domains: segmentation losses, inner loop and outer loop losses, adversarial losses and perceptual losses. Therefore, minimizing the total loss corresponds to solving the optimization problem to look for $Enc^{\star}$ and $F^{\star}$:
\begin{align}
    \label{eq:full_loss}
Enc^{\star},F^{\star}, (Dec^{\star}) =\underset{Enc,F,Dec}{\arg\min}\mathcal{L}_{LoopDA}
\end{align}
where $\mathcal{L}_{LoopDA} = \sum \lambda_{i}\mathcal{L}_{i}$ is a weighted sum, and $i$ stands for the specific loss from Sec.~\ref{sec:loopda} and Sec.~\ref{sec:self_train} by name.

\section{Experiment and Analysis}
\subsection{Datasets and implementation details}
\noindent\textbf{Cityscapes}~\cite{cordts2016cityscapes} is adopted as our labelled daytime source domain. For our domain adaptation task we take its training set containing 2975 pixel-wisely annotated images taken in urban scenes with 19 categories. The original image resolution is 2048×1024.\\ 
\noindent\textbf{Dark Zurich}~\cite{sakaridis2019guided} is considered as our unlabelled nighttime target domain, whose training set contains unlabelled images captured under daytime, twilight and nighttime conditions with a resolution of 1920×1080. To keep consistent with previous papers~\cite{xu2021cdada,wu2021dannet}, we take the 2416 unlabelled night-day image pairs for training. There are also 201 nighttime images with pixel-wise annotation from Dark Zurich dataset, which are separated into validation set (50 images) and test set (151 images), respectively, but the ground-truth of the latter is not publicly available. Evaluation results on the test set can be attained by submitting the model predictions to the provided website by~\cite{sakaridis2019guided}.\\
\noindent\textbf{Nighttime Driving}~\cite{dai2018dark} contains 50 nighttime images with the same resolution as Dark Zurich, which are pixel-wisely annotated following Cityscapes label format. We also evaluate LoopDA based on this dataset.\\
Our implementation of LoopDA is based on Pytorch~\cite{NEURIPS2019_9015} on an NVIDIA Quadro RTX 8000 with 48 GB memory. We use ImageNet~\cite{deng2009imagenet} pretrained {\tt ResNet-101}~\cite{he2016deep} backbone as feature encoder $E$ and adopt {\tt PSPNet}~\cite{chen2017deeplab} for semantic segmentation. During training, our first downscale the original input by a factor of 2 and take $512\times 496$ crops for both domains and set batch size to 2. We use the SGD~\cite{robbins1951stochastic} optimizer with a default learning rate of $2.5\times 10^{-3}$, momentum 0.9, and weight decay
$5\times 10^{-4}$ to train our segmentation network. We set the LOR threshold $\tau$ to 0.5, and set $\lambda_{seg}$ = 1.25, $\lambda_{inner}$ = $\lambda_{outer}$ = 1, $\lambda_{adv}$ = 0.1,  $\lambda_{percep}$ = 1, and in Equation~\ref{eq:PC} we set $\lambda_{z}$ = 0.1 and $\lambda_{l}$ = 0.25.

\subsection{Model evaluation}
\label{sec:evaluation}
Following~\cite{wu2021dannet}, we train LoopDA using PSPNet as the segmentation network, and evaluate our model by submitting the results to the online evaluation site of Dark Zurich~\cite{sakaridis2019guided}. As a common practice, we use mean intersection over union (mIoU) as the final evaluation metric and also IoU for each class. As can be observed in Table~\ref{tab:citytodarkzurich}, comparing with the state-of-the-art, our model attains $46.8$ mIoU, outperforming DANNet by 1.6. Furthermore, we identify that a new milestone can be further reached by having an extra knowledge distillation training stage using {\tt ResNet-101}~\cite{he2016deep} pretrained backbone based on {\tt SimCLRv2}~\cite{chen2020big} following~\cite{zhang2021prototypical}. This brings the best score of LoopDA to 50.6 mIoU. Our model is leading at segmenting road, sidewalk, car, train, bike as well as challenging classes at nighttime such as building, vegetation and sky. Some visual examples are given in Fig.~\ref{fig:visual compare}.\\
To draw similar conclusion, we also provide our model results evaluated on Dark Zurich validation set and Nighttime Driving test set in Table~\ref{tab:val_citytodarkzurich}, where LoopDA also demonstrates impressive results, achieving $37.6$ and $49.6$ mIoU, respectively.
After an extra distillation stage, the results get boosted to $38.7$ and $54.0$ mIoU.
\begin{table*}[htb!]
    \centering
    \setlength{\belowcaptionskip}{-5pt}
    \small
    \setlength{\tabcolsep}{3.5pt}
    \fontsize{7}{11}\selectfont
    \begin{tabular}{c|c|ccccccccccccccccccc|c}
    \hline Method & Arch & \rotatebox{90}{road} & \rotatebox{90}{sdwk} & \rotatebox{90}{bldng} & \rotatebox{90}{wall} & \rotatebox{90}{fence} & \rotatebox{90}{pole} & \rotatebox{90}{light} & \rotatebox{90}{sign} & \rotatebox{90}{veg} & \rotatebox{90}{trrn} & \rotatebox{90}{sky} & \rotatebox{90}{psn} & \rotatebox{90}{rider} & \rotatebox{90}{car} & \rotatebox{90}{truck} & \rotatebox{90}{bus} & \rotatebox{90}{train} & \rotatebox{90}{moto} & \rotatebox{90}{bike} & mIoU  \\ 
    \hline
    \\[-1.6em]
    Deeplabv2~\cite{chen2017deeplab} &$D$& 79.0 & 21.8 & 53.0& 13.3& 11.2& 22.5& 20.2& 22.1& 43.5& 10.4& 18.0& 37.4& 33.8& 64.1&  6.4&  0.0& 52.3& 30.4& 7.4&28.8 \\
    \\[-1.8em]
    RefineNet~\cite{lin2017refinenet} &$R$&  68.8&  23.2& 46.8& 20.8& 12.6& 29.8& 30.4& 26.9& 43.1& 14.3& 0.3& 36.9& 49.7& 63.6&  6.8&  0.2& 24.0& 33.6& 9.3& 28.5\\
    \\[-1.8em]
    PSPNet~\cite{zhao2017pyramid} &$P$&  78.2&  19.0& 51.2& 15.5& 10.6& 30.3& 28.9& 22.0& 56.7& 13.3& 20.8& 38.2& 21.8& 52.1&  1.6&  0.0& 53.2& 23.2& 10.7& 28.8\\
    \\[-1.8em]
    AdaptSegnet~\cite{tsai2018learning}  &$D$&  86.1&  44.2& 55.1& 22.2& 4.8& 21.1& 5.6& 16.7& 37.2& 8.4& 1.2& 35.9& 26.7& 68.2&  45.1&  0.0& 50.1& 33.9& 15.6& 30.4\\
    \\[-1.8em]
    BDL~\cite{li2019bidirectional} &$D$&  85.3&  41.1& 61.9& 32.7& 17.4& 20.6& 11.4& 21.3& 29.4& 8.9& 1.1& 37.4& 22.1& 63.2&  28.2&  0.0& 47.7& {\bf39.4}& 15.7&30.8 \\
    \\[-1.8em]
    DMAda~\cite{dai2018dark} &$R$&  75.9&  29.1& 48.6& 21.3& 14.3& 34.3& 36.8& 29.9& 49.4& 13.8& 0.4& 43.3& 50.2& 69.4&  18.4&  0.0& 27.6& 34.9& 11.9&32.1 \\
    \\[-1.8em]
    GCMA~\cite{sakaridis2019guided} &$R$&  81.7&  46.9& 58.8& 22.0& 20.0& 41.2& {\bf40.5}& {\bf41.6}& 64.8& 31.0& 32.1& {\bf53.5}& 47.5& 75.5&  39.2&  0.0& 49.6& 30.7& 21.0& 42.0\\
    \\[-1.8em]
    MGCDA~\cite{sakaridis2020map} &$R$&  80.3&  49.3& 66.2& 7.8& 11.0& 41.4& 38.9& 39.0& 64.1& 18.0& 55.8&52.1& {\bf53.5}& 74.7&  {\bf66.0}&  0.0& 37.5& 29.1& 22.7 & 42.5\\
    \\[-1.8em]
    CDAda~\cite{xu2021cdada}&$R$&  90.5&  60.6& 67.9& 37.0& 19.3& {\bf42.9}& 36.4& 35.3& 66.9& 24.4& 79.8& 45.4& 42.9& 70.8& 51.7& 0.0& 29.7& 27.7& 26.2& 45.0\\
    \\[-1.8em]
    DANNet~\cite{wu2021dannet} &$P$&  90.4&  60.1& 71.0& 33.6&{\bf22.9}& 30.6& 34.3& 33.7& 70.5& {\bf31.8}& 80.2& 45.7& 41.6& 67.4&  16.8&  0.0& 73.0& 31.6& 22.9& 45.2\\
    \\[-1.8em]
    \hline
    LoopDA(ours) &$P$&  86.3&  46.3& 76.1& 30.3&22.5& 32.5& 34.1& 34.8& 62.6& 19.5& 84.3& 46.6& 51.5& 73.2& 60.7&  {\bf3.1}&  73.4& 26.2& 24.8& 46.8\\
    \\[-1.8em]
    LoopDA$^{\ddagger}(\mathrm{ours})$ &$P$&  {\bf92.1}&  {\bf63.3}& {\bf80.3}& {\bf41.1}&  13.9& 40.8& 39.7& 41.1& {\bf71.3}& 28.4& {\bf85.5}& 50.2& 38.5& {\bf78.2}& 58.5& 3.0 &{\bf77.2} & 26.5& {\bf31.0}&{\bf50.6}\\ 
    \hline
\end{tabular}
\caption{Cityscapes-to-Dark Zurich adaptation results evaluated on the test set. We compare the performance of LoopDA with state-of-the-art methods. In all tables of Sec.~\ref{sec:evaluation}, bold stands for {\bf best}. Regarding network architectures for semantic segmentation: `$D$' stands for Deeplabv2,`$R$' stands for RefineNet and `$P$' stands for PSPNet. $\ddagger$ means an extra distillation stage on nighttime domain using pretrained {\tt ResNet-101}~\cite{he2016deep} based on {\tt SimCLRv2}~\cite{chen2020big} as backbone feature extractor.}
\label{tab:citytodarkzurich}
\end{table*}
\begin{figure*}[htb!]
\centering
\setlength{\belowcaptionskip}{-5pt}
\includegraphics[width=1.6\columnwidth,clip=true,trim=0 0 0 0]{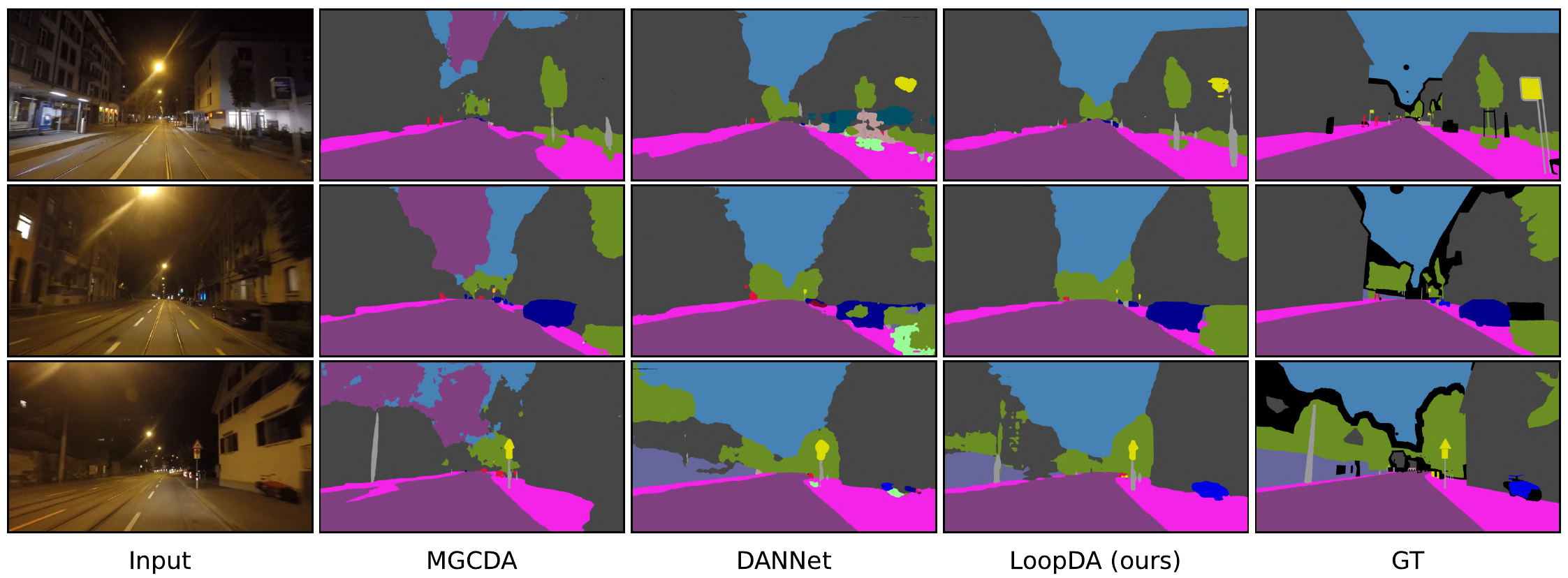}
\caption{Qualitative comparison with state-of-the-art methods for Cityscapes-to-Dark Zurich adaptation on Dark Zurich validation set.}
\label{fig:visual compare}
\end{figure*}
\begin{table}[htb!]
    \centering
    \small
    \setlength{\belowcaptionskip}{-15pt}
    \setlength{\tabcolsep}{4.5pt}
    \fontsize{7}{11}\selectfont
    \begin{tabular}{c|c|cc}
    \hline Method & Arch & Dark Zurich & Night Driving  \\ 
    \hline
    \\[-1.6em]
    AdaptSegnet~\cite{tsai2018learning}  &$D$&  20.2&  34.5 \\
    \\[-1.8em]
    DMAda~\cite{dai2018dark} &$R$&  -&  36.1 \\
    \\[-1.8em]
    GCMA~\cite{sakaridis2019guided} &$R$&  26.7&  45.6\\
    \\[-1.8em]
    MGCDA~\cite{sakaridis2020map} &$R$&  26.1&  49.4\\
    \\[-1.8em]
    CDAda~\cite{xu2021cdada}&$R$&  36.0&  50.9\\
    \\[-1.8em]
    DANNet~\cite{wu2021dannet} &$P$&  36.8&  47.7\\
    \\[-1.8em]
    \hline
    LoopDA(ours) &$P$&  37.6&  49.6\\
    \\[-1.8em]
    LoopDA$^{\ddagger}(\mathrm{ours})$ &$P$&  {\bf38.7}&  {\bf54.0}\\ 
    \hline
\end{tabular}
\caption{mIoU comparison of Cityscapes-to-Dark Zurich adaptation results evaluated on Dark Zurich validation set and Night Driving test set, respectively. Regarding network architectures for semantic segmentation: `$D$' stands for Deeplabv2,`$R$' stands for RefineNet and `$P$' stands for PSPNet. $\ddagger$ means an extra {\tt SimCLRv2}~\cite{chen2020big} distillation stage on nighttime domain.}
\label{tab:val_citytodarkzurich}
\end{table}

\subsection{Ablation study}
\label{sec:ablation}
\begin{table}[htb!]
    \centering
    \small
    \setlength{\belowcaptionskip}{-20pt}
    \setlength{\tabcolsep}{3.0pt}
    \fontsize{7}{11}\selectfont   
    \begin{tabular}{c|c|cc}
    \hline
    Phase & Components  &   mIoU &   $\Delta$     \\ \hline
    \multicolumn{1}{c|}{\multirow{5}{*}{\begin{turn}{270}Warm-up stage\end{turn}}}&
    (\romannum{1}).baseline~\cite{tsai2018learning} on PSPNet ($\mathcal{L}^{d}_{seg}$)&   20.6  &   \textcolor{green}{+}0.0   \\
     &(\romannum{2}). no outer loop (w/o $\mathcal{L}_{percep}$, $\mathcal{L}_{adv}$, $\mathcal{L}^{d2n}_{seg}$, $\mathcal{L}_{outer}$)     &   22.0 &   \textcolor{green}{+}1.4    \\
    &(\romannum{3}). no inner loop (w/o $\mathcal{L}_{inner})$   &   26.3  &   \textcolor{green}{+}5.7  \\
    &(\romannum{4}). no semantic rendering layers  &   28.6  &   \textcolor{green}{+}8.0   \\
    &(\romannum{5}). LoopDA warm-up model     &   29.5&   \textcolor{green}{+}8.9     \\[0.25em]\hline
    \multicolumn{1}{c|}{\multirow{3}{*}{\begin{turn}{270}ST stage\end{turn}}}&(\romannum{6}). no `DNA' (without $\hat{y}_{\scriptsize DNA}$ in $\hat{\mathcal{L}}^{n}_{seg}$)          &   35.7 &   \textcolor{green}{+}15.1   \\
    &(\romannum{7}). LoopDA full configuration ($\hat{\mathcal{L}}^{n}_{seg}$)        &   37.6  &   \textcolor{green}{+}17.0   \\
    &(\romannum{8}). with extra distillation stage (LoopDA$^{\ddagger}$)    &   {\bf 38.7}  &   \textcolor{green}{+}{\textbf18.1}   \\ [0.5em]\hline
    \end{tabular}
    \caption{ Ablation study for Cityscapes-to-Dark Zurich adaptation results evaluated on Dark Zurich validation set.}
\label{tab:ablationstudy}
\end{table}
To better understand how each component of LoopDA affects the final result, in Table~\ref{tab:ablationstudy} we conduct an ablation study on Dark Zurich validation set. We take AdaptSegnet~\cite{tsai2018learning} in row(\romannum{1}) as our baseline approach. Comparing row(\romannum{1}) and row(\romannum{2}), we confirm that adding our inner loop reconstruction only together with semantic rendering can help refine the nighttime predictions, and this brings a 1.4 mIoU performance gain over baseline. This also reveals that the outer loop plays a crucial role in LoopDA training, without which the performance drops from 29.5 (see row(\romannum{5})) to 22.0 mIoU.  Row(\romannum{3}) and row(\romannum{5}) indicate that inner loop is also a dispensable part in LoopDA, and training the warm-up stage without the inner loop results in a performance decrease by 3.2 mIoU. Row(\romannum{4}) shows the result when no semantic rendering layer is involved in training, meaning that the output from segmentation head $F$ cannot get refined with the self-loops. This results in a performance drop by 0.9 mIoU. Row(\romannum{6}) reveals the impact of our proposed `DNA' online label generation strategy. If there is no $\hat{y}_{\scriptsize DNA}$, the result of the self-training stage decreases from 37.6 (row(\romannum{7})) to 35.7 mIoU. Row(\romannum{8}) suggests that an extra {\tt SimCLRv2}~\cite{chen2020big} knowledge distillation stage helps with further improvement, reaching a mIoU of 38.7. More detailed ablation of this part is given in Supplementary. \\
We also provide ablative analysis in Table~\ref{tab:tau_mIoU} to examine the impact of different $\tau$ values for LOR threshold in self-training stage. Setting too low $\tau$ values (e.g., 0.0 or 0.25) means that the self-training mainly relies on $\hat{y}^{ud}_{static}$, which is problematic owing to the camera pose shift between day reference and night inputs. As shown in Fig.~\ref{fig:overfi}, compared to $p^{n}$ based on our proposal, the overfitting issue rises in $p^{n}_{overfit}$ when $\hat{y}^{ud}_{static}$ is dominant. This hinders the model to learn further from pseudo-labels, obtaining 35.7 and 36.4 mIoU, respectively. However, higher $\tau$ values (e.g., 0.75 or 1.0) make $\hat{y}_{\scriptsize DNA}$ play stronger role, which leads to less sufficient self-training (37.1 and 36.9 mIoU). This verifies our selection of $\tau$ (to be 0.5) in the paper, and indicates that our `DNA' self-training strategy is a meaningful solution in dealing with the pose shifted static labels.  
\begin{table}
    \setlength{\belowcaptionskip}{-20pt}
    \centering
    \setlength{\tabcolsep}{6.0pt}
    \fontsize{8}{12}\selectfont
    \begin{tabular}{c|ccccc}
        \hline
        $\tau$              &  $0.0$   &   $0.25$   &     $0.5$   &       $0.75$   &   $1.0$                      \\ \hline\hline
        mIoU    & 35.7    & 36.4 & 37.6      & 37.1           & 36.9       
       \\ \hline
    \end{tabular}
    \caption{mIoU comparison of applying different $\tau$ values for online label generation during self-training stage. Results obtained on Dark Zurich validation set.}
    \label{tab:tau_mIoU}
\end{table}

\begin{figure}[t!]
\centering
\setlength{\belowcaptionskip}{-12pt}
\includegraphics[width=1.0\columnwidth,clip=true,trim=0 0 0 0]{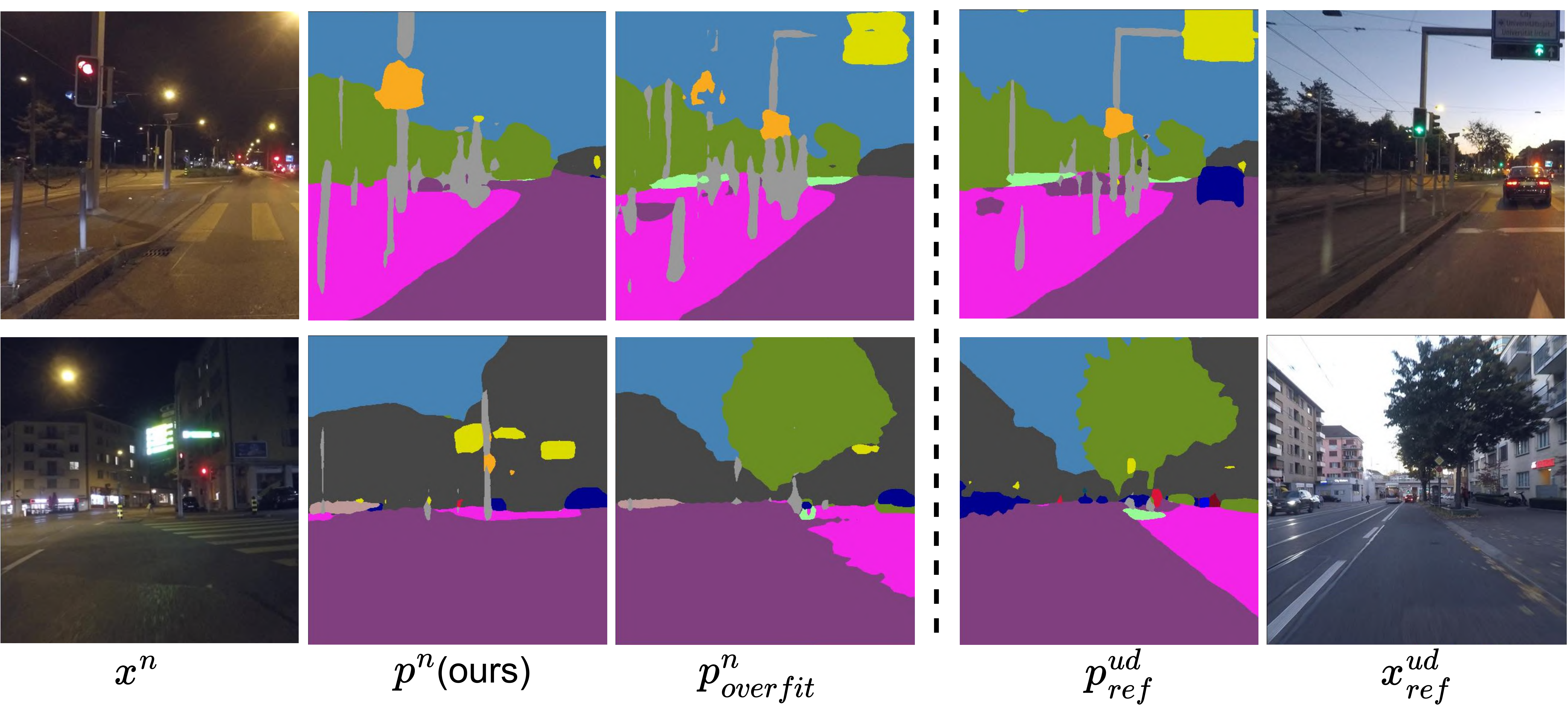}
\caption{Visual comparison of self-training results without introducing our $\hat{y}_{\scriptsize DNA}$. $p^{n}_{overfit}$ gets overfitted on daytime reference static label $\hat{y}^{ud}_{static}$ due to the day-night camera pose shift.}
\label{fig:overfi}
\end{figure}

\subsection{Discussion}
\begin{figure}[t!]
\centering
\setlength{\belowcaptionskip}{-12pt}
\includegraphics[width=0.7\columnwidth,clip=true,trim=0 0 0 0]{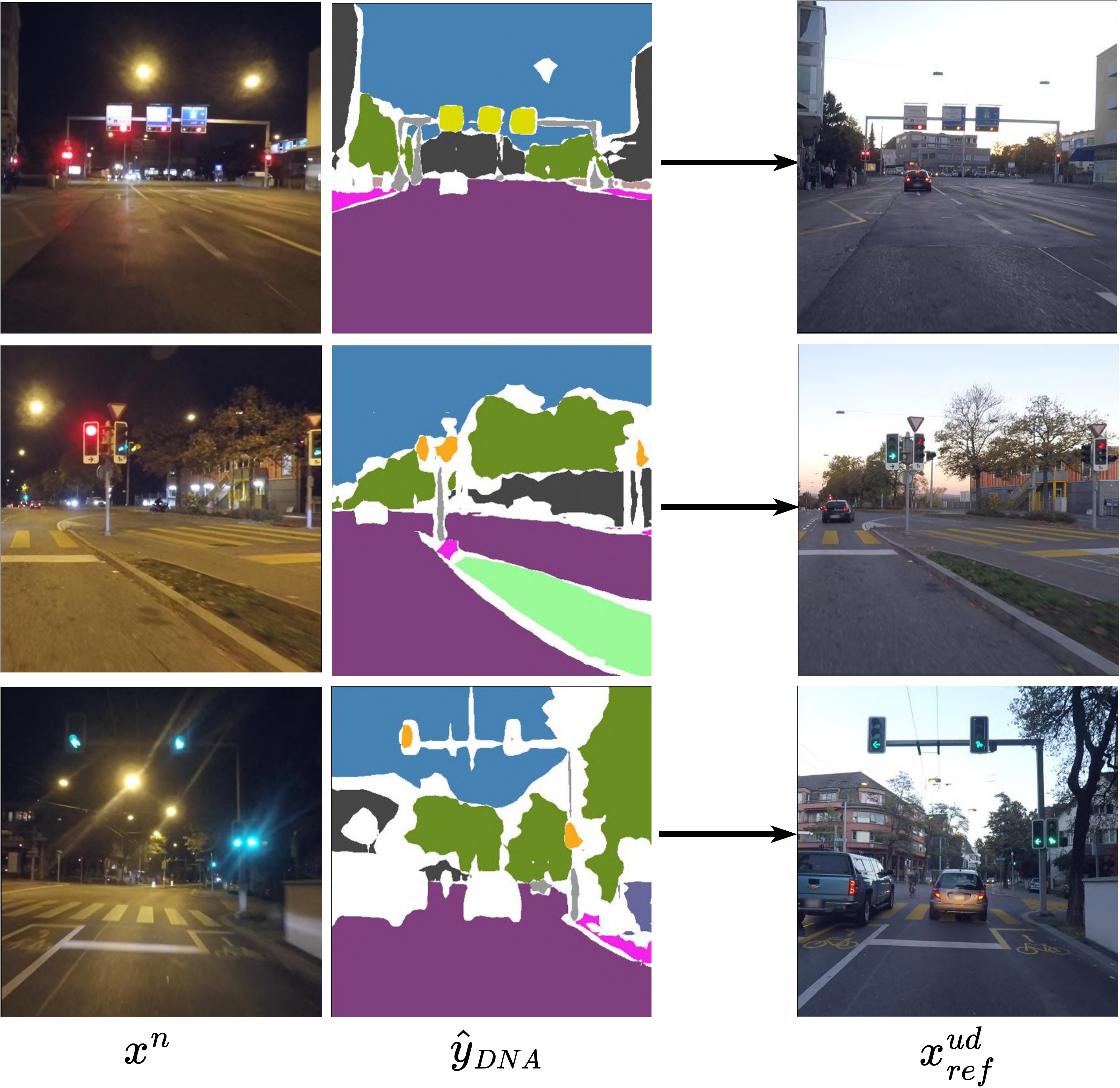}
\caption{Examples of retrieved day-night training pairs based on our $\hat{y}_{\scriptsize DNA}$, satisfying condition $LOR\ge 0.5$. }
\label{fig:retrieval}
\end{figure}

\noindent\textbf{Limitations}. Although our LoopDA framework demonstrates impressive performances on benchmark datasets for domain adaptive nighttime segmentation, we found that there is still space to improve. First of all, in choosing between $\hat{y}_{\scriptsize DNA}$ and $\hat{y}^{ud}_{static}$ for self-training, we rely on SSC, however, there can also be close day-night image pairs where SSC do not exist. In this case, the static labels are not fully utilized because $\hat{y}^{ud}_{static}$ is dismissed. Secondly, for the task of image translation, there are powerful GAN architectures that ensure higher fidelity outputs, but require much more memory for training. Therefore, a trade-off between GAN image quality and memory consumption should be made. Furthermore, as a basic proof of concept, we follow~\cite{li2019bidirectional} for offline pseudo-label generation, which can be replaced by more advanced pseudo-labelling techniques in the field of domain adaptation for better mIoU. These aspects will be further explored as future work.\\     
\noindent\textbf{Chances}. We point out that there exists potential of extending $\hat{y}_{\scriptsize DNA}$ obtained from UDA to other research areas. For instance, one possible direction can be image retrieval. Day-night image retrieval has been an open challenge since acquiring domain-robust descriptors for this scenario is difficult, let alone the problematic dynamic objects that lead to mismatches. However, we conduct a simple experiment in Fig.~\ref{fig:retrieval} by saving day-night image pairs retrieved based on our $\hat{y}_{\scriptsize DNA}$ with $LOR\ge 0.5$. Interestingly, we observe that these obtained cross-domain image pairs are captured almost at the same location. To this point, we argue that using larger $\tau$ to check LOR between cross-domain image pairs can provide a new approach for image retrieval. A further application can be 3D reconstruction given a cross-domain dataset, where similar scenes can be better matched using $\hat{y}_{\scriptsize DNA}$.       
\section{Conclusion}
In this work, we propose LoopDA for domain adaptive nighttime image segmentation. It constructs different levels of self-loops, starting with the inputs and ending up with an aim to recover the inputs from the encoded latent features rendered by the segmentation outputs. In this way, the intra-domain inner loop enables the semantic predictions to be refined, and the inter-domain outer loop enforces the learned cross-domain knowledge to be better aligned. For self-training, to alleviate the misguidance of pose-shifted static labels produced by the daytime reference images, we propose a co-teaching mechanism. It allows an offline signal, together with an online `DNA' signal that checks the agreed day-night predictions for pseudo-supervision on nighttime domain. Our self-training strategy makes direct use of the network outputs without introducing extra operation modules, which is easy to ensemble for other similar tasks. The efficacy of LoopDA is verified through extensive experiments on benchmark datasets, attaining state-of-the-art performance for adapting nighttime semantic segmentation.

{\small
\bibliographystyle{ieee_fullname}
\bibliography{loopda}

\begin{thebibliography}{10}\itemsep=-1pt

\bibitem{arjovsky2017wasserstein}
Martin Arjovsky, Soumith Chintala, and L{\'e}on Bottou.
\newblock Wasserstein generative adversarial networks.
\newblock In {\em International conference on machine learning}, pages
  214--223. PMLR, 2017.

\bibitem{baldi2012autoencoders}
Pierre Baldi.
\newblock Autoencoders, unsupervised learning, and deep architectures.
\newblock In {\em Proceedings of ICML workshop on unsupervised and transfer
  learning}, pages 37--49. JMLR Workshop and Conference Proceedings, 2012.

\bibitem{bank2020autoencoders}
Dor Bank, Noam Koenigstein, and Raja Giryes.
\newblock Autoencoders.
\newblock {\em arXiv preprint arXiv:2003.05991}, 2020.

\bibitem{ben2007analysis}
Shai Ben-David, John Blitzer, Koby Crammer, Fernando Pereira, et~al.
\newblock Analysis of representations for domain adaptation.
\newblock {\em Advances in neural information processing systems}, 19:137,
  2007.

\bibitem{chang2019all}
Wei-Lun Chang, Hui-Po Wang, Wen-Hsiao Peng, and Wei-Chen Chiu.
\newblock All about structure: Adapting structural information across domains
  for boosting semantic segmentation.
\newblock In {\em Proceedings of the IEEE/CVF Conference on Computer Vision and
  Pattern Recognition}, pages 1900--1909, 2019.

\bibitem{chen2017deeplab}
Liang-Chieh Chen, George Papandreou, Iasonas Kokkinos, Kevin Murphy, and Alan~L
  Yuille.
\newblock Deeplab: Semantic image segmentation with deep convolutional nets,
  atrous convolution, and fully connected crfs.
\newblock {\em IEEE transactions on pattern analysis and machine intelligence},
  40(4):834--848, 2017.

\bibitem{chen2020big}
Ting Chen, Simon Kornblith, Kevin Swersky, Mohammad Norouzi, and Geoffrey~E
  Hinton.
\newblock Big self-supervised models are strong semi-supervised learners.
\newblock {\em Advances in neural information processing systems},
  33:22243--22255, 2020.

\bibitem{choi2019self}
Jaehoon Choi, Taekyung Kim, and Changick Kim.
\newblock Self-ensembling with gan-based data augmentation for domain
  adaptation in semantic segmentation.
\newblock In {\em Proceedings of the IEEE/CVF International Conference on
  Computer Vision}, pages 6830--6840, 2019.

\bibitem{cordts2016cityscapes}
Marius Cordts, Mohamed Omran, Sebastian Ramos, Timo Rehfeld, Markus Enzweiler,
  Rodrigo Benenson, Uwe Franke, Stefan Roth, and Bernt Schiele.
\newblock The cityscapes dataset for semantic urban scene understanding.
\newblock In {\em Proceedings of the IEEE conference on computer vision and
  pattern recognition}, pages 3213--3223, 2016.

\bibitem{dai2018dark}
Dengxin Dai and Luc Van~Gool.
\newblock Dark model adaptation: Semantic image segmentation from daytime to
  nighttime.
\newblock In {\em 2018 21st International Conference on Intelligent
  Transportation Systems (ITSC)}, pages 3819--3824. IEEE, 2018.

\bibitem{dalal2005histograms}
Navneet Dalal and Bill Triggs.
\newblock Histograms of oriented gradients for human detection.
\newblock In {\em 2005 IEEE computer society conference on computer vision and
  pattern recognition (CVPR'05)}, volume~1, pages 886--893. Ieee, 2005.

\bibitem{daume2009frustratingly}
Hal Daum{\'e}~III.
\newblock Frustratingly easy domain adaptation.
\newblock {\em arXiv preprint arXiv:0907.1815}, 2009.

\bibitem{deng2009imagenet}
Jia Deng, Wei Dong, Richard Socher, Li-Jia Li, Kai Li, and Li Fei-Fei.
\newblock Imagenet: A large-scale hierarchical image database.
\newblock In {\em 2009 IEEE conference on computer vision and pattern
  recognition}, pages 248--255. Ieee, 2009.

\bibitem{du2019ssf}
Liang Du, Jingang Tan, Hongye Yang, Jianfeng Feng, Xiangyang Xue, Qibao Zheng,
  Xiaoqing Ye, and Xiaolin Zhang.
\newblock Ssf-dan: Separated semantic feature based domain adaptation network
  for semantic segmentation.
\newblock In {\em Proceedings of the IEEE/CVF International Conference on
  Computer Vision}, pages 982--991, 2019.

\bibitem{feng2020deep}
Di Feng, Christian Haase-Sch{\"u}tz, Lars Rosenbaum, Heinz Hertlein, Claudius
  Glaeser, Fabian Timm, Werner Wiesbeck, and Klaus Dietmayer.
\newblock Deep multi-modal object detection and semantic segmentation for
  autonomous driving: Datasets, methods, and challenges.
\newblock {\em IEEE Transactions on Intelligent Transportation Systems},
  22(3):1341--1360, 2020.

\bibitem{ganin2015unsupervised}
Yaroslav Ganin and Victor Lempitsky.
\newblock Unsupervised domain adaptation by backpropagation.
\newblock In {\em International conference on machine learning}, pages
  1180--1189. PMLR, 2015.

\bibitem{godard2019digging}
Cl{\'e}ment Godard, Oisin Mac~Aodha, Michael Firman, and Gabriel~J Brostow.
\newblock Digging into self-supervised monocular depth estimation.
\newblock In {\em Proceedings of the IEEE/CVF International Conference on
  Computer Vision}, pages 3828--3838, 2019.

\bibitem{goodfellow2014generative}
Ian Goodfellow, Jean Pouget-Abadie, Mehdi Mirza, Bing Xu, David Warde-Farley,
  Sherjil Ozair, Aaron Courville, and Yoshua Bengio.
\newblock Generative adversarial nets.
\newblock In Z. Ghahramani, M. Welling, C. Cortes, N. Lawrence, and K.~Q.
  Weinberger, editors, {\em Advances in Neural Information Processing Systems},
  volume~27, 2014.

\bibitem{hao2020labelenc}
Miao Hao, Yitao Liu, Xiangyu Zhang, and Jian Sun.
\newblock Labelenc: A new intermediate supervision method for object detection.
\newblock In {\em European Conference on Computer Vision}, pages 529--545.
  Springer, 2020.

\bibitem{he2021masked}
Kaiming He, Xinlei Chen, Saining Xie, Yanghao Li, Piotr Doll{\'a}r, and Ross
  Girshick.
\newblock Masked autoencoders are scalable vision learners.
\newblock {\em arXiv preprint arXiv:2111.06377}, 2021.

\bibitem{he2016deep}
Kaiming He, Xiangyu Zhang, Shaoqing Ren, and Jian Sun.
\newblock Deep residual learning for image recognition.
\newblock In {\em Proceedings of the IEEE conference on computer vision and
  pattern recognition}, pages 770--778, 2016.

\bibitem{hoffman2018cycada}
Judy Hoffman, Eric Tzeng, Taesung Park, Jun-Yan Zhu, Phillip Isola, Kate
  Saenko, Alexei Efros, and Trevor Darrell.
\newblock Cycada: Cycle-consistent adversarial domain adaptation.
\newblock In {\em International conference on machine learning}, pages
  1989--1998. PMLR, 2018.

\bibitem{huang2018multimodal}
Xun Huang, Ming-Yu Liu, Serge Belongie, and Jan Kautz.
\newblock Multimodal unsupervised image-to-image translation.
\newblock In {\em Proceedings of the European conference on computer vision
  (ECCV)}, pages 172--189, 2018.

\bibitem{jolicoeur2018relativistic}
Alexia Jolicoeur-Martineau.
\newblock The relativistic discriminator: a key element missing from standard
  {GAN}.
\newblock In {\em International Conference on Learning Representations}, 2019.

\bibitem{karras2019style}
Tero Karras, Samuli Laine, and Timo Aila.
\newblock A style-based generator architecture for generative adversarial
  networks.
\newblock In {\em Proceedings of the IEEE/CVF Conference on Computer Vision and
  Pattern Recognition}, pages 4401--4410, 2019.

\bibitem{kim2020learning}
Myeongjin Kim and Hyeran Byun.
\newblock Learning texture invariant representation for domain adaptation of
  semantic segmentation.
\newblock In {\em Proceedings of the IEEE/CVF Conference on Computer Vision and
  Pattern Recognition}, pages 12975--12984, 2020.

\bibitem{kingma2013auto}
Diederik~P Kingma and Max Welling.
\newblock Auto-encoding variational bayes.
\newblock {\em arXiv preprint arXiv:1312.6114}, 2013.

\bibitem{krizhevsky2012imagenet}
Alex Krizhevsky, Ilya Sutskever, and Geoffrey~E Hinton.
\newblock Imagenet classification with deep convolutional neural networks.
\newblock {\em Advances in neural information processing systems},
  25:1097--1105, 2012.

\bibitem{lecun2015deep}
Yann LeCun, Yoshua Bengio, and Geoffrey Hinton.
\newblock Deep learning.
\newblock {\em nature}, 521(7553):436--444, 2015.

\bibitem{lecun1998gradient}
Yann LeCun, L{\'e}on Bottou, Yoshua Bengio, and Patrick Haffner.
\newblock Gradient-based learning applied to document recognition.
\newblock {\em Proceedings of the IEEE}, 86(11):2278--2324, 1998.

\bibitem{li2019bidirectional}
Yunsheng Li, Lu Yuan, and Nuno Vasconcelos.
\newblock Bidirectional learning for domain adaptation of semantic
  segmentation.
\newblock In {\em Proceedings of the IEEE/CVF Conference on Computer Vision and
  Pattern Recognition}, pages 6936--6945, 2019.

\bibitem{lin2017refinenet}
Guosheng Lin, Anton Milan, Chunhua Shen, and Ian Reid.
\newblock Refinenet: Multi-path refinement networks for high-resolution
  semantic segmentation.
\newblock In {\em Proceedings of the IEEE conference on computer vision and
  pattern recognition}, pages 1925--1934, 2017.

\bibitem{liu2017unsupervised}
Ming-Yu Liu, Thomas Breuel, and Jan Kautz.
\newblock Unsupervised image-to-image translation networks.
\newblock In {\em Proceedings of the 31st International Conference on Neural
  Information Processing Systems}, NIPS'17, pages 700--708, Red Hook, NY, USA,
  2017.

\bibitem{long2015fully}
Jonathan Long, Evan Shelhamer, and Trevor Darrell.
\newblock Fully convolutional networks for semantic segmentation.
\newblock In {\em Proceedings of the IEEE conference on computer vision and
  pattern recognition}, pages 3431--3440, 2015.

\bibitem{long2016unsupervised}
Mingsheng Long, Han Zhu, Jianmin Wang, and Michael~I Jordan.
\newblock Unsupervised domain adaptation with residual transfer networks.
\newblock {\em arXiv preprint arXiv:1602.04433}, 2016.

\bibitem{mao2017least}
Xudong Mao, Qing Li, Haoran Xie, Raymond~YK Lau, Zhen Wang, and Stephen
  Paul~Smolley.
\newblock Least squares generative adversarial networks.
\newblock In {\em Proceedings of the IEEE international conference on computer
  vision}, pages 2794--2802, 2017.

\bibitem{pan2020unsupervised}
Fei Pan, Inkyu Shin, Francois Rameau, Seokju Lee, and In~So Kweon.
\newblock Unsupervised intra-domain adaptation for semantic segmentation
  through self-supervision.
\newblock In {\em Proceedings of the IEEE/CVF Conference on Computer Vision and
  Pattern Recognition}, pages 3764--3773, 2020.

\bibitem{park2019semantic}
Taesung Park, Ming-Yu Liu, Ting-Chun Wang, and Jun-Yan Zhu.
\newblock Semantic image synthesis with spatially-adaptive normalization.
\newblock In {\em Proceedings of the IEEE/CVF Conference on Computer Vision and
  Pattern Recognition}, pages 2337--2346, 2019.

\bibitem{NEURIPS2019_9015}
Adam Paszke, Sam Gross, Francisco Massa, Adam Lerer, James Bradbury, Gregory
  Chanan, Trevor Killeen, Zeming Lin, Natalia Gimelshein, Luca Antiga, et~al.
\newblock Pytorch: An imperative style, high-performance deep learning library.
\newblock {\em Advances in neural information processing systems}, 32, 2019.

\bibitem{robbins1951stochastic}
Herbert Robbins and Sutton Monro.
\newblock A stochastic approximation method.
\newblock {\em The annals of mathematical statistics}, pages 400--407, 1951.

\bibitem{romera2019bridging}
Eduardo Romera, Luis~M Bergasa, Kailun Yang, Jose~M Alvarez, and Rafael Barea.
\newblock Bridging the day and night domain gap for semantic segmentation.
\newblock In {\em 2019 IEEE Intelligent Vehicles Symposium (IV)}, pages
  1312--1318. IEEE, 2019.

\bibitem{rumelhart1986learning}
David~E Rumelhart, Geoffrey~E Hinton, and Ronald~J Williams.
\newblock Learning representations by back-propagating errors.
\newblock {\em nature}, 323(6088):533--536, 1986.

\bibitem{saito2018maximum}
Kuniaki Saito, Kohei Watanabe, Yoshitaka Ushiku, and Tatsuya Harada.
\newblock Maximum classifier discrepancy for unsupervised domain adaptation.
\newblock In {\em Proceedings of the IEEE conference on computer vision and
  pattern recognition}, pages 3723--3732, 2018.

\bibitem{sakaridis2019guided}
Christos Sakaridis, Dengxin Dai, and Luc~Van Gool.
\newblock Guided curriculum model adaptation and uncertainty-aware evaluation
  for semantic nighttime image segmentation.
\newblock In {\em Proceedings of the IEEE/CVF International Conference on
  Computer Vision}, pages 7374--7383, 2019.

\bibitem{sakaridis2020map}
Christos Sakaridis, Dengxin Dai, and Luc Van~Gool.
\newblock Map-guided curriculum domain adaptation and uncertainty-aware
  evaluation for semantic nighttime image segmentation.
\newblock {\em arXiv preprint arXiv:2005.14553}, 2020.

\bibitem{sankaranarayanan2018learning}
Swami Sankaranarayanan, Yogesh Balaji, Arpit Jain, Ser~Nam Lim, and Rama
  Chellappa.
\newblock Learning from synthetic data: Addressing domain shift for semantic
  segmentation.
\newblock In {\em Proceedings of the IEEE Conference on Computer Vision and
  Pattern Recognition}, pages 3752--3761, 2018.

\bibitem{shen2021tridentadapt}
Fengyi Shen, Akhil Gurram, Ahmet~Faruk Tuna, Onay Urfalioglu, and Alois Knoll.
\newblock Tridentadapt: Learning domain-invariance via source-target
  confrontation and self-induced cross-domain augmentation.
\newblock {\em arXiv preprint arXiv:2111.15300}, 2021.

\bibitem{sun2019see}
Lei Sun, Kaiwei Wang, Kailun Yang, and Kaite Xiang.
\newblock See clearer at night: towards robust nighttime semantic segmentation
  through day-night image conversion.
\newblock In {\em Artificial Intelligence and Machine Learning in Defense
  Applications}, volume 11169, page 111690A. International Society for Optics
  and Photonics, 2019.

\bibitem{tsai2018learning}
Yi-Hsuan Tsai, Wei-Chih Hung, Samuel Schulter, Kihyuk Sohn, Ming-Hsuan Yang,
  and Manmohan Chandraker.
\newblock Learning to adapt structured output space for semantic segmentation.
\newblock In {\em Proceedings of the IEEE conference on computer vision and
  pattern recognition}, pages 7472--7481, 2018.

\bibitem{wei2021masked}
Chen Wei, Haoqi Fan, Saining Xie, Chao-Yuan Wu, Alan Yuille, and Christoph
  Feichtenhofer.
\newblock Masked feature prediction for self-supervised visual pre-training.
\newblock {\em arXiv preprint arXiv:2112.09133}, 2021.

\bibitem{wu2021dannet}
Xinyi Wu, Zhenyao Wu, Hao Guo, Lili Ju, and Song Wang.
\newblock Dannet: A one-stage domain adaptation network for unsupervised
  nighttime semantic segmentation.
\newblock In {\em Proceedings of the IEEE/CVF Conference on Computer Vision and
  Pattern Recognition}, pages 15769--15778, 2021.

\bibitem{xu2021cdada}
Qi Xu, Yinan Ma, Jing Wu, Chengnian Long, and Xiaolin Huang.
\newblock Cdada: A curriculum domain adaptation for nighttime semantic
  segmentation.
\newblock In {\em Proceedings of the IEEE/CVF International Conference on
  Computer Vision}, pages 2962--2971, 2021.

\bibitem{yang2020label}
Jinyu Yang, Weizhi An, Sheng Wang, Xinliang Zhu, Chaochao Yan, and Junzhou
  Huang.
\newblock Label-driven reconstruction for domain adaptation in semantic
  segmentation.
\newblock In {\em European Conference on Computer Vision}, pages 480--498.
  Springer, 2020.

\bibitem{yue2019domain}
Xiangyu Yue, Yang Zhang, Sicheng Zhao, Alberto Sangiovanni-Vincentelli, Kurt
  Keutzer, and Boqing Gong.
\newblock Domain randomization and pyramid consistency: Simulation-to-real
  generalization without accessing target domain data.
\newblock In {\em Proceedings of the IEEE/CVF International Conference on
  Computer Vision}, pages 2100--2110, 2019.

\bibitem{zhang2020cross}
Pan Zhang, Bo Zhang, Dong Chen, Lu Yuan, and Fang Wen.
\newblock Cross-domain correspondence learning for exemplar-based image
  translation.
\newblock In {\em Proceedings of the IEEE/CVF Conference on Computer Vision and
  Pattern Recognition}, pages 5143--5153, 2020.

\bibitem{zhang2021prototypical}
Pan Zhang, Bo Zhang, Ting Zhang, Dong Chen, Yong Wang, and Fang Wen.
\newblock Prototypical pseudo label denoising and target structure learning for
  domain adaptive semantic segmentation.
\newblock In {\em Proceedings of the IEEE/CVF Conference on Computer Vision and
  Pattern Recognition}, pages 12414--12424, 2021.

\bibitem{zhang2018unreasonable}
Richard Zhang, Phillip Isola, Alexei~A Efros, Eli Shechtman, and Oliver Wang.
\newblock The unreasonable effectiveness of deep features as a perceptual
  metric.
\newblock In {\em Proceedings of the IEEE conference on computer vision and
  pattern recognition}, pages 586--595, 2018.

\bibitem{zhao2017pyramid}
Hengshuang Zhao, Jianping Shi, Xiaojuan Qi, Xiaogang Wang, and Jiaya Jia.
\newblock Pyramid scene parsing network.
\newblock In {\em Proceedings of the IEEE conference on computer vision and
  pattern recognition}, pages 2881--2890, 2017.

\bibitem{zhu2017unpaired}
Jun-Yan Zhu, Taesung Park, Phillip Isola, and Alexei~A Efros.
\newblock Unpaired image-to-image translation using cycle-consistent
  adversarial networks.
\newblock In {\em Proceedings of the IEEE international conference on computer
  vision}, pages 2223--2232, 2017.

\bibitem{zhu2018penalizing}
Xinge Zhu, Hui Zhou, Ceyuan Yang, Jianping Shi, and Dahua Lin.
\newblock Penalizing top performers: Conservative loss for semantic
  segmentation adaptation.
\newblock In {\em Proceedings of the European Conference on Computer Vision
  (ECCV)}, pages 568--583, 2018.

\bibitem{zou2018unsupervised}
Yang Zou, Zhiding Yu, BVK Kumar, and Jinsong Wang.
\newblock Unsupervised domain adaptation for semantic segmentation via
  class-balanced self-training.
\newblock In {\em Proceedings of the European conference on computer vision
  (ECCV)}, pages 289--305, 2018.

\bibitem{zou2019confidence}
Yang Zou, Zhiding Yu, Xiaofeng Liu, BVK Kumar, and Jinsong Wang.
\newblock Confidence regularized self-training.
\newblock In {\em Proceedings of the IEEE/CVF International Conference on
  Computer Vision}, pages 5982--5991, 2019.

\end{thebibliography}
}

\end{document}